\documentclass[letterpaper, 10 pt, conference]{ieeeconf} 

\IEEEoverridecommandlockouts

\usepackage{graphics} 
\usepackage{epsfig} 
\usepackage{mathptmx} 
\usepackage{times} 
\usepackage{amsmath} 
\usepackage{amssymb}  
\usepackage{multirow}
\usepackage{tabularx} 
\usepackage{adjustbox}
\usepackage{graphicx}
\usepackage{subcaption}
\usepackage{float}
\usepackage{caption}

\title{\LARGE \bf
TS3IM: Unveiling Structural Similarity in Time Series through Image Similarity Assessment Insights
}

\author{Yuhan Liu$^{1*}$ and Ke Tu$^{2}$
\thanks{$^{1}$Software Engineering Institute, East China Normal University, Shanghai, China,
        {\tt\small urmeas917@gmail.com}}%
\thanks{$^{2}$Software Engineering Institute, East China Normal University, Shanghai, China}%
}

\begin{document}

\maketitle
\thispagestyle{empty}
\pagestyle{empty}

\begin{abstract}

In the realm of time series analysis, accurately measuring similarity is crucial for applications such as forecasting, anomaly detection, and clustering. However, existing metrics often fail to capture the complex, multidimensional nature of time series data, limiting their effectiveness and application. This paper introduces the Structured Similarity Index Measure for Time Series (TS3IM), a novel approach inspired by the success of the Structural Similarity Index Measure (SSIM) in image analysis, tailored to address these limitations by assessing structural similarity in time series. TS3IM evaluates multiple dimensions of similarity—trend, variability, and structural integrity—offering a more nuanced and comprehensive measure. This metric represents a significant leap forward, providing a robust tool for analyzing temporal data and offering more accurate and comprehensive sequence analysis and decision support in fields such as monitoring power consumption, analyzing traffic flow, and adversarial recognition. Our extensive experimental results also show that compared with traditional methods that rely heavily on computational correlation, TS3IM is 1.87 times more similar to Dynamic Time Warping (DTW) in evaluation results and improves by more than 50\% in adversarial recognition.

\end{abstract}

\section{Introduction}
Calculating the similarity between two time series is a basic operation that supports numerous search and analysis activities for time series. Despite the existence of multiple metrics to capture the resemblance, an intuitive and universally accepted metric for the degree of similarity is clearly lacking, especially compared to the diversity of methods available in the image and signal processing fields. This gap highlights the need for a new approach to assessing structural similarity in time series data, aiming to capture the complex dynamics and patterns inherent in time series.
In the realm of time series analysis, methodologies for assessing similarity have historically been centered around specific aspects of the data, such as quantifying the distance between sequences. But this also highlights the limitations of current methods, which often fail to provide an intuitive explanation of sequence similarity, thus hindering their usefulness in tasks such as anomaly detection and decision-making. While the cross-correlation function \cite{vorburger2011applications} (CCF) is one of the few tools that provide an intuitive explanation of sequence similarity by considering the correlation between two series, it fails short in representing structural similarities in time series and is not sensitive to extreme values.

\begin{figure}[htbp]
\centering
\includegraphics[scale=0.3]{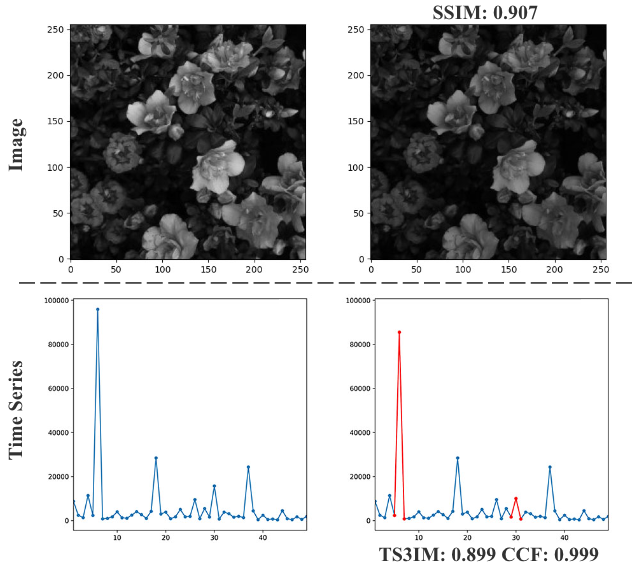} 
\caption{The upper line is an example of SSIM. Except for contrast, almost all other structural information of the two pictures is the same, but SSIM still captured this information and reflected it in the score. Subsequently, the following line delineates instances of the time series field. The sequence changes values at 2 time steps and one of them changes drastically, but it can be seen from the results that the existing relative similarity method CCF cannot accurately capture the structural nuances of the sequence, while TS3IM is able to.}
      \label{figure-intro}
\end{figure}

Navigating through the complexities of time series similarity measurements, we encounter critical gaps in the methodology that prompt us to examine two fundamental challenges. \textbf{The key challenges we explore in this context are twofold: (1) How can we measure the structural similarity in time series data analogous to methods used in image analysis? (2) How does the application of such a comprehensive metric for time series similarity measurement compare to existing methods in downstream tasks? Is it more effective?}
To address the first challenge, we introduce the TS3IM, a novel metric designed to quantify structural similarities between time series datasets.  Drawing inspiration from the \cite{wang2004image} SSIM in image analysis, TS3IM evaluates the trend, variability, and structural integrity of time series data, offering a multidimensional perspective on similarity.  This approach not only fills the existing gap in time series analysis but also provides a robust framework for capturing complex patterns and dynamics that are indicative of real-world phenomena. As shown in Fig. \ref{figure-intro}, we clearly show TS3IM’s superiority over traditional methods like CCF in the time series field. This advantage mirrors that of SSIM, enabling the evaluation of multiple dimensions of time series, offering a more nuanced and comprehensive measure.

Addressing the second challenge, we first validate the TS3IM's relevance and effectiveness through a Correlation Analysis with DTW, establishing the metric's rationality and applicability.  Further, comprehensive Similarity Evaluations conducted across nine diverse datasets demonstrate TS3IM's superior performance in identifying similarities within time series data.  Additionally, our analysis reveals TS3IM's enhanced capability to detect adversarial samples compared to existing methods, showcasing its potential in securing time series data against sophisticated attacks.
In summary, our main contributions of this work are threefold:
\begin{itemize}
    \item We propose TS3IM, an innovative metric for assessing structural similarity in time series data, inspired by the success of SSIM in image quality assessment. 
    \item Through rigorous validation, including Correlation Analysis with DTW, we demonstrate TS3IM's effectiveness, sensitivity to changes in time series, and its ability to reflect the overall integrity of the data, emphasizing its application in practical data analysis tasks.
    \item We illustrate TS3IM's unique advantage in identifying adversarial samples within time series data, highlighting its potential to enhance data security and integrity in sensitive applications.
\end{itemize}

\section{Related Work}
\subsection{Image Similarity Measurements}
In image analysis, the endeavor to measure image similarity has spawned various methods, each targeting different visual aspects. Early metrics like Mean Squared Error (MSE) and Peak Signal-to-Noise Ratio (PSNR) focus on pixel discrepancies, providing direct quantitative assessments yet often misaligning with human perception. As advancements ensued, emphasis shifted towards sophisticated techniques valuing perceptual image features. The Feature Similarity Index \cite{zhang2011fsim} (FSIM) and Gradient Magnitude Similarity Deviation \cite{xue2013gradient} (GMSD) exemplify this, with FSIM evaluating feature-based similarity mimicking human vision, and GMSD examining structural integrity via gradient magnitudes. Additionally, the Perceptual Hash Algorithm (pHash) abstracts an image's essence into a hash format, enabling similarity detection across transformations. Amid these developments, the SSIM distinguishes itself by holistically assessing luminance, contrast, and structure, offering a perceptually attuned image similarity measure. SSIM's comprehensive approach has established it as a benchmark in image quality assessment. This array of image analysis methodologies underscores a void in time series analysis, lacking a comparably nuanced, perception-aligned metric. The TS3IM fills this gap, inspired by SSIM's success, to chart a novel course in time series similarity measurement. TS3IM's introduction aims to address this disparity, promising a more detailed approach to evaluating temporal data similarity.

\subsection{Time Series Similarity Measurements}
In the landscape of time series analysis, several metrics have been developed to measure similarity, each tailored for specific aspects of the data. DTW stands out for its ability to match sequences with temporal shifts, offering flexibility unmatched by traditional metrics like ED, which calculates the straightforward distance between corresponding points in two sequences. Beyond these, the CCF identifies linear relationships across different lags, whereas metrics like Longest Common Subsequence (LCSS) and Edit Distance on Real Sequence (EDR) delve into pattern matching and sequences' structural alterations, respectively. More novel methods such as NeuTS focus on using machine learning methods to measure the similarity between time series \cite{yao2020linear}. Despite the utility of these metrics in their respective domains, from signal processing to pattern recognition, they often fall short in capturing the full spectrum of similarity in time series data and fail to give a generally accepted measure of similarity, be it due to computational constraints, sensitivity to noise, or the inability to account for multidimensional similarity aspects. This is where the TS3IM comes into play, advancing beyond traditional metrics by evaluating multiple dimensions of similarity, including trend, variability, and structural integrity. TS3IM's comprehensive approach not only addresses the limitations of prior metrics but also enhances the robustness and applicability of similarity assessments in diverse applications, setting a new benchmark in the field.

\section{Approach Design}

\begin{figure*}[htbp]
\centering
\includegraphics[scale=0.25]{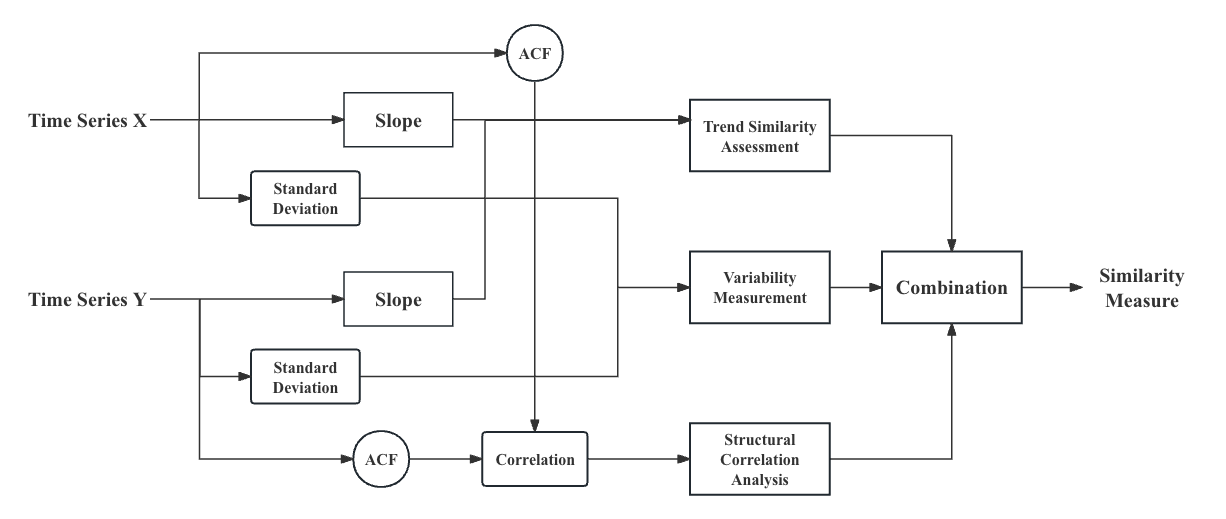} 
\caption{Diagram of the TS3IM measurement system.}
\label{figure-label}
\end{figure*}

The construction of the TS3IM metric unfolds methodically across three critical phases, namely Trend Similarity Assessment, Variability Measurement, and Structural Correlation Analysis, each delving into a distinct aspect of time series similarity. The system diagram is shown in Fig. \ref{figure-label}. 

\subsection{Trend Similarity Assessment}
The first step in constructing the TS3IM metric involves assessing the overall trends in time series data by comparing their linear regression slopes. This approach conceptualizes time series as one-dimensional images, where values indicate intensity levels, akin to luminance comparison in SSIM. The slope of a linear regression on a time series reflects its trend or the general direction of the data over time. The calculation is given by:

\begin{equation}
\mu_X = \frac{\sum_{i=1}^{m} (i - \bar{i})(X_i - \bar{X})}{\sum_{i=1}^{m} (i - \bar{i})^2}.
\end{equation}

The slope of a linear regression is utilized to quantify the trend similarity between time series. Significant fluctuations near extreme values in a series can lead to deviations in the slope from its expected behavior, causing changes in similarity scores. This variation enhances the sensitivity of TS3IM to minor changes, enabling it to capture even subtle differences in adversarial environments.

\subsection{Variability Measurement}

In the second step, the contrast of time series data is evaluated by calculating the overall variance, which measures the variability or fluctuation within the series. This step corresponds to the contrast comparison in SSIM, where the standard deviation is used to quantify signal variation. For time series, the variance is given as follows:

\begin{equation}
\sigma_X = \sqrt{\frac{1}{m} \sum_{i=1}^{m}(X_i - \bar{X})^2}.
\end{equation}

This variance reflects the degree of signal variation within the time series, providing a measure of its contrast.

\subsection{Structural Correlation Analysis}
The third step involves comparing the structure of time series by evaluating their autocorrelation functions \cite{mestre2021functional} (ACF). This comparison is analogous to SSIM's structure comparison, where the correlation coefficient between two signals is calculated. For time series, the structural similarity is assessed by:

\begin{equation}
\sigma_{XY} = \frac{\sum_{i=1}^{m}(ACF_{X_i} \cdot ACF_{Y_i})}{\sqrt{\sum_{i=1}^{m}(ACF_{X_i})^2 \cdot \sum_{i=1}^{m}(ACF_{Y_i})^2}}.
\end{equation}

This step measures the structural similarity between two time series, focusing on their internal patterns and dynamics.

\subsection{Comprehensive Similarity Index}

The TS3IM metric integrates the assessments from the above steps into a comprehensive similarity index. This is achieved through a formula that equally weighs the luminance (trend), contrast (variability), and structure (correlation) aspects of the time series:

\begin{equation}
\Theta(X, Y) = \frac{(2\mu_X\mu_Y + C_1)(2\sigma_{XY} + C_2)}{(\mu_X^2 + \mu_Y^2 + C_1)(\sigma_X^2 + \sigma_Y^2 + C_2)},
\end{equation}
Constants \(C_1\) and \(C_2\) are introduced to ensure numerical stability, particularly when the denominator approaches zero. These constants are defined as:

\begin{equation}
C_1 = (K_1L_X)^2, \quad C_2 = (K_2L_Y)^2,
\end{equation}
Here, $L_X$ and $L_Y$ represent the maximum values of time series $X$ and $Y$, respectively. $K_1 \ll 1$ and $K_2 \ll 1$ are small constants. To ensure that the final result represents the index measure, we normalize the $C(X, Y)$ by dividing it by value when the two time series are identical. Notice that the formula $(2\sigma_{XY} + C_2)/ 2(\sigma_{X}^2 + C_2)$ represents the same value of $C(X,Y)$ when the time series $X$ and $Y$ are identical. Finally, to ensure the TS3IM metric falls within a [0,1] range and accurately reflects the similarity index, the final formula is normalized and adjusted if necessary:

\begin{equation}
\text{TS3IM}(X, Y) = \frac{2\Theta(X, Y)(\sigma_{X}^2 + C_2)}{2\sigma_{XY} + C_2}.
\end{equation}

This structured approach to the TS3IM metric's construction offers a novel method for assessing time series similarity, drawing inspiration from proven techniques in image quality evaluation.
    
\section{Experimental Results}
We are interested in three key properties of TS3IM: its representation of time series correlation aligns with the trend observed in distance measurements, it offers a more comprehensive evaluation of time series similarity compared to the CCF, and its effectiveness in the domain of identifying adversarial samples.

\begin{table*}[h]
\caption{A summary of the benchmark datasets}
\centering
\begin{adjustbox}{width=\textwidth}
\begin{tabular}{c|ccccccccc}
\hline
\textbf{Datasets} & \textbf{Electricity} & \textbf{Traffic} & \textbf{BirdChicken} & \textbf{ECGFiveDays} & \textbf{Lightning2} & \textbf{FE} & \textbf{PJMW} & \textbf{NonInvasiveFetalECGThorax1} & \textbf{NonInvasiveFetalECGThorax2} \\ \hline
\textbf{Length}   & 321                  & 862              & 512                  & 861                  & 637                 & 366         & 366           & 750                                 & 750                                 \\
\textbf{Count}    & 26304                & 17544            & 20                   & 136                  & 61                  & 8           & 17            & 1965                                & 1965                                \\ \hline
\end{tabular}
\end{adjustbox}
\label{tab_dataset}
\end{table*}


\subsection{Experiment Setup}

\noindent\textbf{Datasets.} To illustrate that TS3IM can be applied to datasets of different scales and multiple domains, we explore several experiments on nine datasets. These datasets include the ElectricityLoadDiagrams20112014 dataset \cite{lichman2013uci}, which contains electricity consumption in kWh recorded every 15 minutes from 2011 to 2014, 48 months (2015-2016) of hourly energy consumption data from PJM in Megawatts \cite{kaggle_energy_dataset}, hourly energy consumption data from the California Department of Transportation \cite{pemsdotca}, and datasets from the UCR time series classification archives \cite{dau2019ucr}, which only contain test sets. The details are given in Table \ref{tab_dataset} (the count row states the size of the dataset).

\noindent\textbf{Comparison Models.} In previous work only CCF was an indicator to measure the relative similarity, but there were several measures of distance between sequences, we selected the most popular DTW and ED to verify that TS3IM is more detailed and comprehensive, so TS3IM are compared against:
\begin{itemize}
    \item CCF: a standard similarity measure for baseline.
    \item DTW: a method that can effectively measure absolute distances between sequences, to highlight the effectiveness of the TS3IM as described in section \uppercase\expandafter{\romannumeral3}.
    \item ED: provides a baseline of the straight-line distance between two data points in Euclidean space for comparison in our experiments.
\end{itemize}
\begin{figure*}[htbp]
    \centering
    \begin{subfigure}[b]{0.45\textwidth}
        \centering
        \includegraphics[scale=0.25]{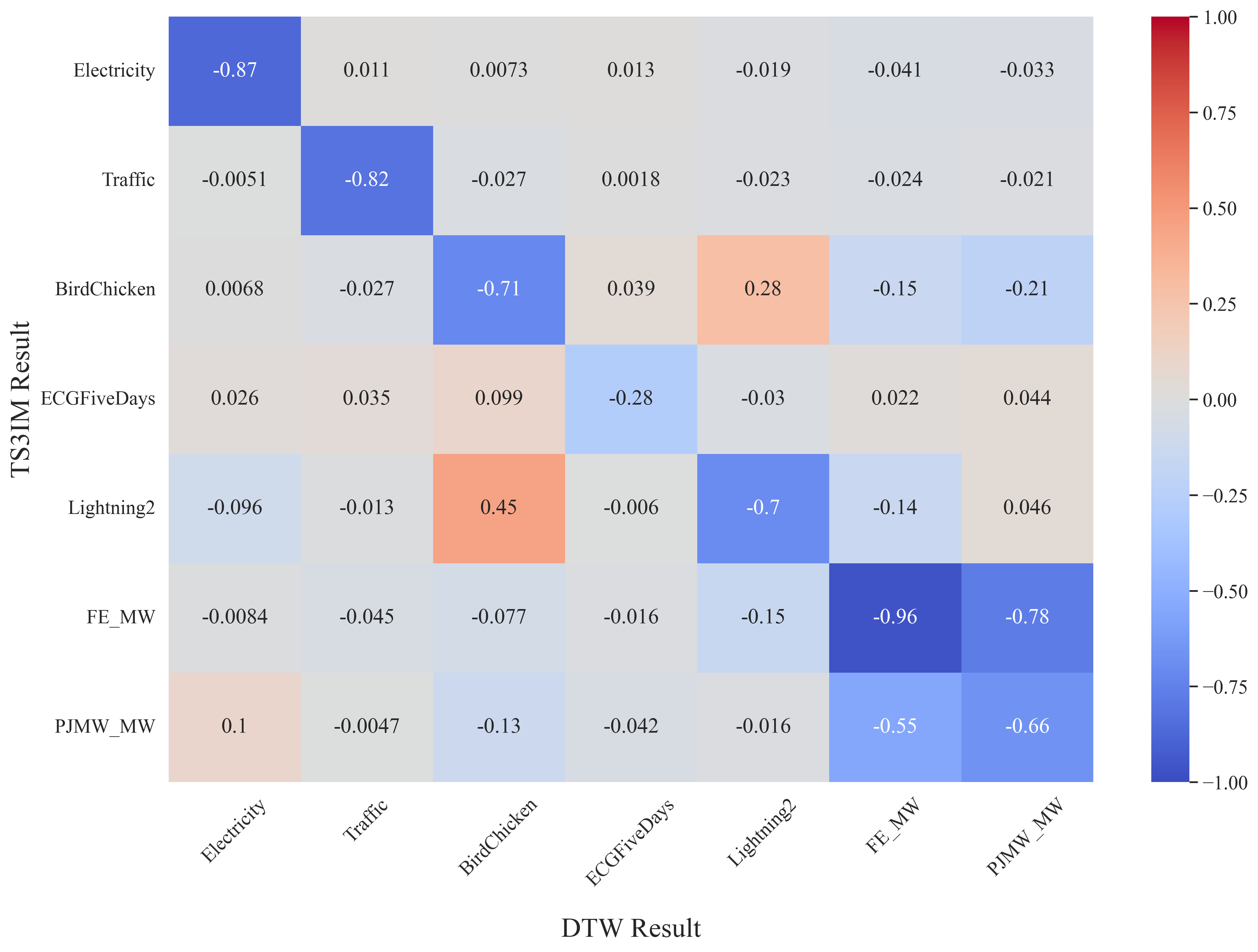}
        \caption{TS3IM-DTW}
        \label{fig-ts3im}
    \end{subfigure}
    \quad
    \hspace{1cm}
    \begin{subfigure}[b]{0.45\textwidth}
        \centering
        \includegraphics[scale=0.25]{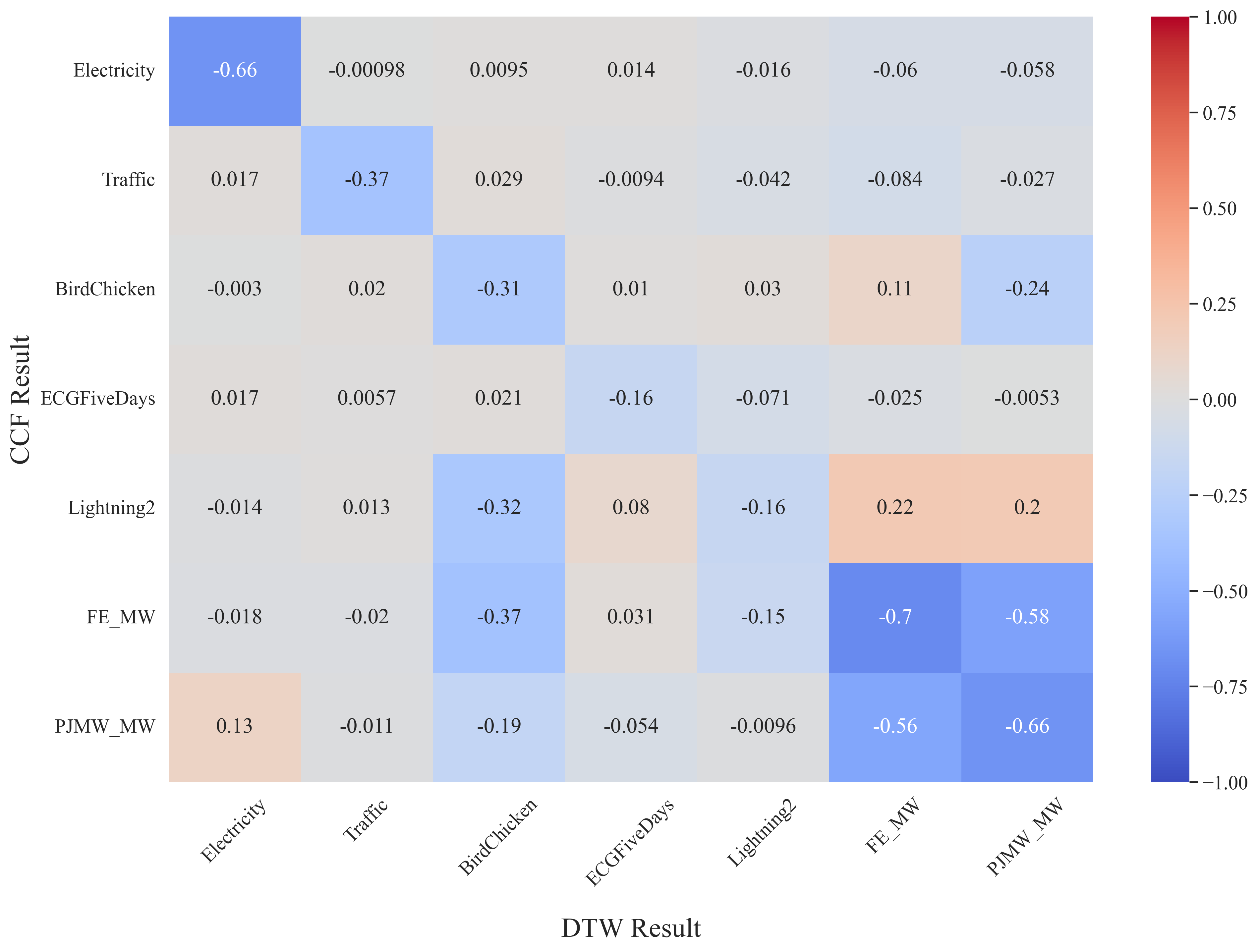}
        \caption{CCF-DTW}
        \label{fig-ccf}
    \end{subfigure}
    \caption{Confusion matrices for the Pearson correlation coefficients: TS3IM-DTW Fig. \ref{fig-ts3im}, CCF-DTW Fig. \ref{fig-ccf}.}
    \label{fig-dtw}
\end{figure*}


\subsection{Correlation Analysis}
First, to determine whether the results of TS3IM accurately reflect the similarity between two sequences, we calculated the correlation between the outcomes of TS3IM and those of existing distance measurement methods. Specifically, we selected the DTW method, which is increasingly recognized as the most effective measure in many domains \cite{li2018review}. We employed Pearson correlation coefficients to quantify the correlation between the results of TS3IM and DTW, as illustrated in Fig. \ref{fig-ts3im}. For comparison, we also calculated the correlation between the results of CCF and DTW, depicted in Fig. \ref{fig-ccf}. Each column corresponds to the results of DTW on these datasets, while each row represents the results of TS3IM and CCF. Note that the CCF values are scaled to be in the same range for ease of comparison. 

The numerical range in the figure spans from -1.00 to 1.00, with values closer to -1 indicating a stronger negative correlation. The average correlation between TS3IM and DTW is -0.71, followed by the correlation between CCF and DTW, which is -0.43, which indicates that TS3IM can effectively reflect the similarity between two sequences better than CCF. 

Both TS3IM and CCF show relatively high correlations with DTW on the FE and PJMW datasets. This phenomenon is likely due to the datasets' short length and the substantial magnitude of changes between sequences. The Euclidean distance averages around $10^7$ for each time step, indicating significant variations. As a result, both TS3IM and CCF are sensitive to these pronounced changes, leading to their higher correlations with DTW on the FE and PJMW datasets. However, on other datasets, TS3IM yields higher similarity to DTW due to its comprehensive consideration of Trend Similarity, Variability, and Structural Correlation. Our comprehensive TS3IM results in a more objective and comprehensive similarity assessment compared to CCF.

\subsection{Comprehensive Evaluation}

\begin{table}[h]
\centering
\captionsetup{justification=centering, labelsep=newline}
\caption{Similarity Values For All Drawn Sample Group Obtained By TS3IM, CCF, DTW And FD}
\label{table_example}
\begin{tabular}{cc|cccc}
\hline
\multicolumn{2}{c|}{\multirow{2}{*}{\textbf{Dataset rows}}} & \multicolumn{4}{c}{\textbf{Similarity Result}}                                                             \\ \cline{3-6} 
\multicolumn{2}{c|}{}                                       & \multicolumn{2}{c|}{\textbf{relative similarity}}  & \multicolumn{2}{c}{\textbf{distance-like similarity}} \\ \hline
\textbf{X}                   & \textbf{Y}                   & \textbf{TS3IM} & \multicolumn{1}{c|}{\textbf{CCF}} & \textbf{DTW}               & \textbf{ED}              \\ \hline
\textbf{500}                 & \textbf{501}                 & 0.975          & \multicolumn{1}{c|}{0.970}        & 4.653                      & 19.546                   \\
\textbf{600}                 & \textbf{601}                 & 0.995          & \multicolumn{1}{c|}{0.979}        & 2.335                      & 7.466                    \\
\textbf{700}                 & \textbf{701}                 & 0.972          & \multicolumn{1}{c|}{0.925}        & 14.183                     & 304.171                  \\
\textbf{800}                 & \textbf{801}                 & 0.854          & \multicolumn{1}{c|}{0.869}        & 26.730                     & 715.43                   \\
\textbf{900}                 & \textbf{901}                 & 0.991          & \multicolumn{1}{c|}{0.892}        & 6.579                      & 4.171                    \\ \hline
\end{tabular}
\end{table}

Motivated by the need for a comprehensive evaluation of time series similarity, TS3IM integrates various aspects of a sequence's characteristics. By calculating slope, standard deviation, and autocorrelation coefficient, provides a more comprehensive view of similarities and can be reflected in the results when the series shows obvious volatility. Arguably, in terms of evaluation factors, TS3IM provides a more robust assessment of time series similarity compared to CCF. Based on the above reasons, we selected the Traffic dataset, which provides a suitable environment for evaluating the performance of similarity methods for our experiments, to demonstrate the effectiveness of TS3IM. We specifically sampled every 100 rows from the 500th to the 900th row to ensure a representative subset for our analysis.
\begin{figure}[htbp]
    \begin{subfigure}[b]{0.4\textwidth}
        \centering
        \includegraphics[scale=0.2]{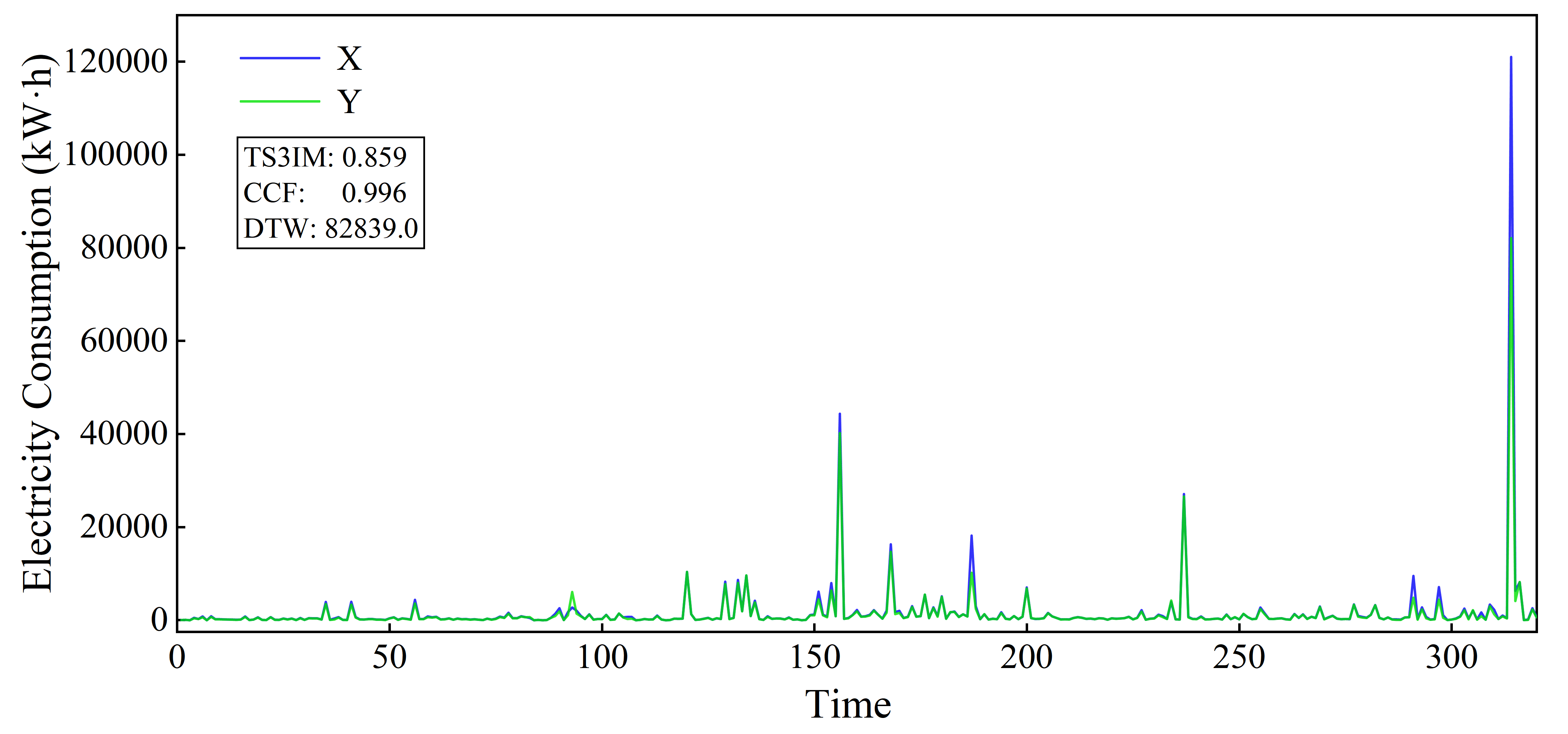}
        \caption{}
        \label{fig-plot1}
    \end{subfigure}
    \quad
    \begin{subfigure}[b]{0.4\textwidth}
        \centering
        \includegraphics[scale=0.2]{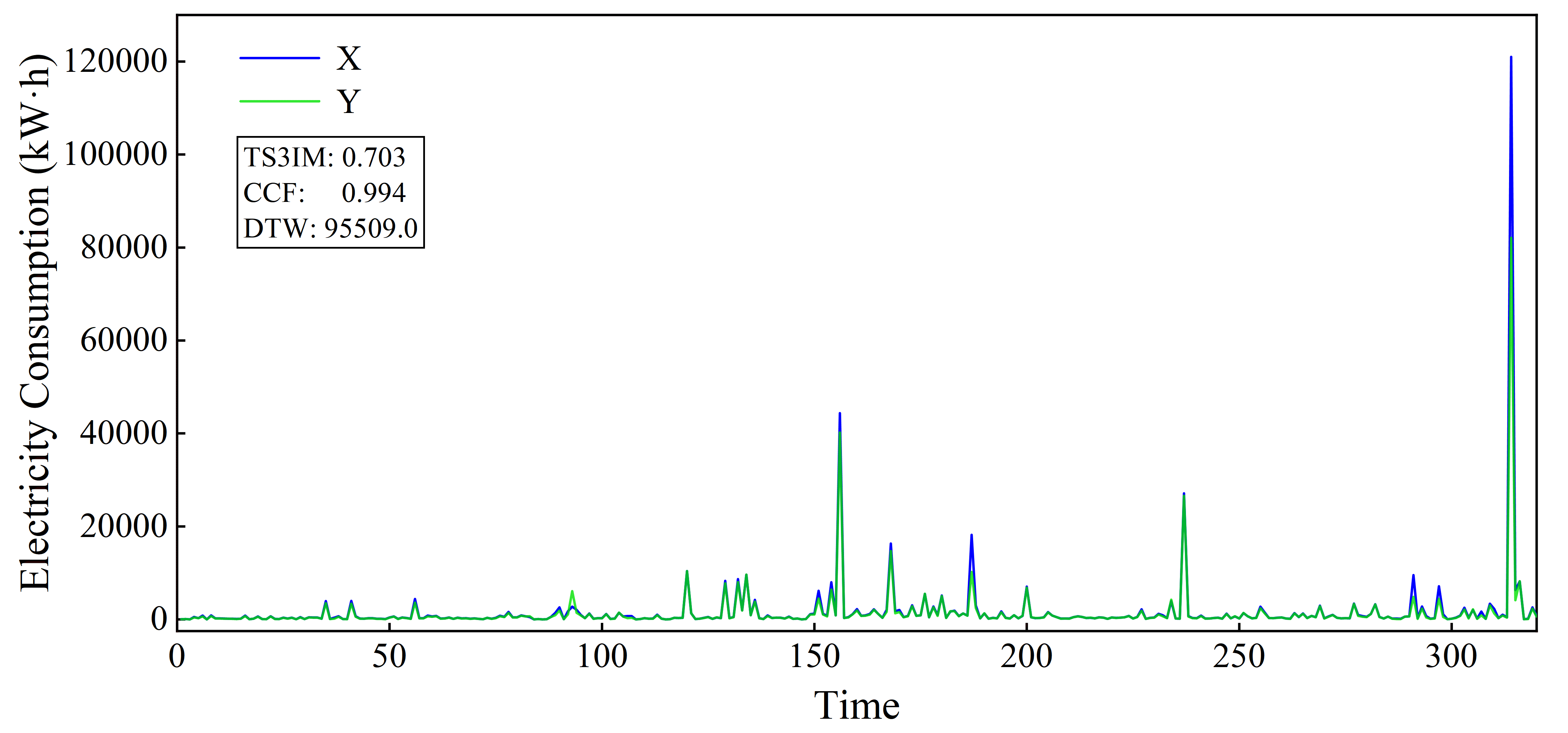}
        \caption{}
        \label{fig-plot2}
    \end{subfigure}
    \caption{Two plots depicting two sets of time series data, each comprising X and Y sequences.}
    \label{fig-subfigures}
\end{figure}

 It can be observed in Table \ref{table_example} that the correlation between the measurements of TS3IM and DTW is higher than that of CCF and DTW. When the DTW value increases, TS3IM decreases. The former means that the distance-like quantity between sequences increases, and the latter means that the sequence similarity decreases. Especially in the last sample, where DTW is relatively small, indicating a small difference between the two sequences, TS3IM calculates a high similarity value close to 1 for the two sequences. The value of CCF is not as high as TS3IM, which is 0.892, and is even smaller than the value of 0.925 when DTW is about 14 in the third row, indicating that it cannot sensitively reflect the similarity between the two sequences. As can be seen from the comparison of the data in the table, TS3IM is more sensitive and comprehensive. These characteristics also give it stronger advantages in time series analysis in various real-world fields.

\subsection{Real-world Application}

We then present two comparisons of time series from the real-world dataset Electricity, which collected hourly electricity consumption data from 321 customers from 2012 to 2014. In Fig. \ref{fig-plot1}, the DTW distance value is 82839.0, with TS3IM and CCF providing similarity scores of 0.859 and 0.996, respectively. In Fig. \ref{fig-plot2}, the DTW value is 95509.0, with TS3IM scoring 0.703 and CCF scoring 0.994. The change in DTW is 12670.0, while in TS3IM it is 0.156 and in CCF it is only 0.002.

\begin{figure}[htbp]
\centering
\includegraphics[scale=0.2]{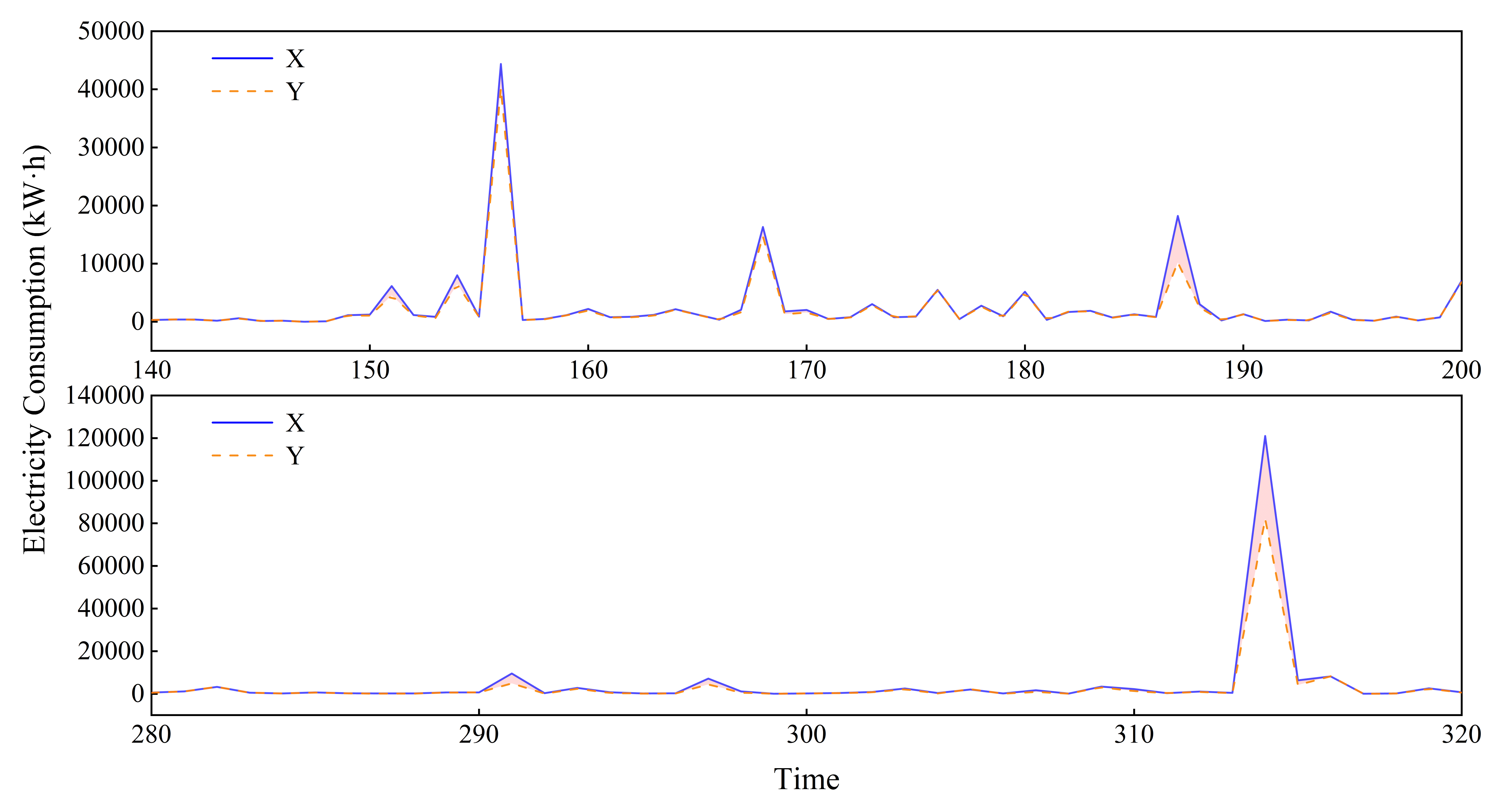} 
\caption{Two subsequence plots from Fig. \ref{fig-plot2}, representing the subsequences at [140, 200] and [250, 320], respectively.}
\label{fig-sub}
\end{figure}

From the comparison of the two plots in Fig. \ref{fig-subfigures}, it can be observed that for the majority of users, there is no significant variation in electricity consumption. Fig. \ref{fig-sub} represents a subsequence of Fig. \ref{fig-plot2}, showing significant differences in electricity consumption for a few individual users. These extreme changes in consumption are causing drastic changes in DTW. Here are the users whose consumption changes exceed 4000 kWh, listed in descending order: Users 312, 185, 289, and 154 experienced consumption changes of 38900 kWh, 8001 kWh, 4676 kWh, and 4166 kWh, respectively. In real life, large changes in consumption from these users can cause imbalances in the load on the electricity grid, affecting its stability. This can lead to voltage fluctuations or even blackouts if not managed effectively. Therefore, timely assessment of significant changes in electricity consumption by individual users is crucial.

The comparison of CCF and TS3IM's results indicates that TS3IM is more sensitive to the changes occurring in these two plots compared to DTW and CCF. TS3IM is capable of capturing these drastic changes in the subsequences and reflecting them in the similarity scores. This capability allows for early detection of issues such as equipment malfunctions or changes in consumer behavior, enabling proactive management and optimization of electricity usage.

\subsection{Adversarial Sample Identification}
Currently, adversarial examples play a crucial role in revealing the robustness of neural networks \cite{carlini2017towards}. However, in the absence of domain experts to discern whether adversarial examples are sufficiently indistinguishable, similarity metrics play a significant role. Unfortunately, the time series domain currently lacks similarity metrics similar to those in the image domain that can quantify the quality of adversarial examples. Our TS3IM approach addresses this gap by providing a comprehensive assessment of time series similarity to distinguish between adversarial examples and real data.

\begin{table*}[]
\centering
\caption{Similarity Detection Result for Adversarial Examples on Different Models and Attacks}
\label{table_attack}
\begin{tabular}{c|c|ccccc|ccccc}
\hline
\multirow{2}{*}{Model}                  & \multirow{2}{*}{\textbf{Attack}} & \multicolumn{5}{c|}{\textbf{NonInvasiveFetalECGThorax1}}                                                                      & \multicolumn{5}{c}{\textbf{NonInvasiveFetalECGThorax2}}                                                                     \\ \cline{3-12} 
                                        &                                  & \textbf{TS3IM}  & \multicolumn{1}{c|}{\textbf{CCF}}    & \multicolumn{1}{l|}{\textbf{TS3IM  / CCF}} & \textbf{DTW} & \textbf{ED} & \textbf{TS3IM} & \multicolumn{1}{c|}{\textbf{CCF}}   & \multicolumn{1}{c|}{\textbf{TS3IM / CCF}} & \textbf{DTW} & \textbf{ED} \\ \hline
\multirow{3}{*}{\textbf{FCN}}           & \textbf{DeepFool}                & 0.015           & \multicolumn{1}{c|}{0.007}           & \multicolumn{1}{c|}{2.1}                & 34.73        & 181.52      & 3.108          & \multicolumn{1}{c|}{0.007}          & \multicolumn{1}{c|}{444}                & 66.18        & 8251.61     \\
                                        & \textbf{FGSM}                    & 0.065           & \multicolumn{1}{c|}{0.035}           & \multicolumn{1}{c|}{1.9}                & 191.03       & 153.64      & 0.064          & \multicolumn{1}{c|}{0.037}          & \multicolumn{1}{c|}{1.7}                & 190.22       & 106.81      \\
                                        & \textbf{PGD}                     & 0.055           & \multicolumn{1}{c|}{0.033}           & \multicolumn{1}{c|}{1.7}                & 182.26       & 251.98      & 0.047          & \multicolumn{1}{c|}{0.032}          & \multicolumn{1}{c|}{1.5}                & 186.52       & 569.78      \\ \hline
\multirow{3}{*}{\textbf{GRU}}           & \textbf{DeepFool}                & 0.110           & \multicolumn{1}{c|}{0.048}           & \multicolumn{1}{c|}{2.3}                & 86.26        & 2706.71     & 0.038          & \multicolumn{1}{c|}{0.019}          & \multicolumn{1}{c|}{2.0}                & 35.51        & 286.64      \\
                                        & \textbf{FGSM}                    & 0.117  & \multicolumn{1}{c|}{0.030}  & \multicolumn{1}{c|}{3.9}       & 131.08       & 2284.11     & 0.117 & \multicolumn{1}{c|}{0.029} & \multicolumn{1}{c|}{4.0}       & 134.62       & 2197.21     \\
                                        & \textbf{PGD}                     & 0.078  & \multicolumn{1}{c|}{0.015}  & \multicolumn{1}{c|}{5.2}       & 71.87        & 1880.52     & 0.090 & \multicolumn{1}{c|}{0.015} & \multicolumn{1}{c|}{6.0}       & 77.65        & 1039.73     \\ \hline
\multirow{3}{*}{\textbf{ResCNN}}        & \textbf{DeepFool}                & 0.0003 & \multicolumn{1}{c|}{0.0001} & \multicolumn{1}{c|}{3.0}     & 6.592        & 0.004       & 0.012          & \multicolumn{1}{c|}{0.006}          & \multicolumn{1}{c|}{2.0}                & 41.63        & 1030.78     \\
                                        & \textbf{FGSM}                    & 0.070           & \multicolumn{1}{c|}{0.035}           & \multicolumn{1}{c|}{2.0}                & 195.45       & 17.28       & 0.063          & \multicolumn{1}{c|}{0.034}          & \multicolumn{1}{c|}{1.9}                & 194.67       & 72.78       \\
                                        & \textbf{PGD}                     & 0.046           & \multicolumn{1}{c|}{0.031}           & \multicolumn{1}{c|}{1.5}                & 172.44       & 534.37      & 0.046          & \multicolumn{1}{c|}{0.028}          & \multicolumn{1}{c|}{1.6}                & 166.71       & 857.90      \\ \hline
\multirow{3}{*}{\textbf{InceptionTime}} & \textbf{DeepFool}                & 0.002           & \multicolumn{1}{c|}{0.001}           & \multicolumn{1}{c|}{2.0}                & 11.25        & 0.54        & 0.003 & \multicolumn{1}{c|}{0.001} & \multicolumn{1}{c|}{3.0}       & 14.62        & 0.22        \\
                                        & \textbf{FGSM}                    & 0.066           & \multicolumn{1}{c|}{0.035}           & \multicolumn{1}{c|}{1.9}                & 192.74       & 57.47       & 0.070          & \multicolumn{1}{c|}{0.033}          & \multicolumn{1}{c|}{2.1}                & 187.96       & 231.74      \\
                                        & \textbf{PGD}                     & 0.051           & \multicolumn{1}{c|}{0.029}           & \multicolumn{1}{c|}{1.8}                & 160.81       & 14.45       & 0.080 & \multicolumn{1}{c|}{0.026} & \multicolumn{1}{c|}{3.0}      & 168.20       & 517.37      \\ \hline
\multirow{3}{*}{\textbf{LSTM}}          & \textbf{DeepFool}                & 0.0002          & \multicolumn{1}{c|}{1.6828}          & \multicolumn{1}{c|}{0.0001}             & 0.53         & 0.63        & 0.0004         & \multicolumn{1}{c|}{3.0545}         & \multicolumn{1}{c|}{0.0001}             & 0.93         & 1.34        \\
                                        & \textbf{FGSM}                    & 0.089  & \multicolumn{1}{c|}{0.030}  & \multicolumn{1}{c|}{3.0}       & 115.67       & 4211.04     & 0.085          & \multicolumn{1}{c|}{0.029}          & \multicolumn{1}{c|}{2.9}                & 122.18       & 4681.04     \\
                                        & \textbf{PGD}                     & 0.051  & \multicolumn{1}{c|}{0.017}  & \multicolumn{1}{c|}{3.0}      & 63.30        & 2365.37     & 0.055 & \multicolumn{1}{c|}{0.014} & \multicolumn{1}{c|}{3.9}      & 66.82        & 2643.71     \\ \hline
\multirow{3}{*}{\textbf{XCM}}           & \textbf{DeepFool}                & 0.020           & \multicolumn{1}{c|}{0.008}           & \multicolumn{1}{c|}{2.5}                & 33.38        & 61.50       & 0.022          & \multicolumn{1}{c|}{0.009}          & \multicolumn{1}{c|}{2.4}                & 36.97        & 197.43      \\
                                        & \textbf{FGSM}                    & 0.070           & \multicolumn{1}{c|}{0.035}           & \multicolumn{1}{c|}{2.0}                & 157.11       & 293.98      & 0.065          & \multicolumn{1}{c|}{0.033}          & \multicolumn{1}{c|}{2.0}                & 159.03       & 709.87      \\
                                        & \textbf{PGD}                     & 0.061           & \multicolumn{1}{c|}{0.030}           & \multicolumn{1}{c|}{2.0}                & 127.04       & 286.89      & 0.044          & \multicolumn{1}{c|}{0.029}          & \multicolumn{1}{c|}{1.5}                & 129.82       & 2078.59     \\ \hline
\end{tabular}
\end{table*}

 In our study, we have considered six neural network models, namely FCN, GRU, ResCNN, InceptionTime, LSTM, and XCM, under three adversarial attack methods for generating attack samples, including DeepFool, FGSM, and PGD. We evaluated the effectiveness of TS3IM, CCF, DTW, and ED in detecting discrepancies between attack samples and original samples from two datasets, namely NonInvasiveFetalECGThorax1 and NonInvasiveFetalECGThorax2.
 
\begin{figure}[H]
\centering
\includegraphics[scale=0.65]{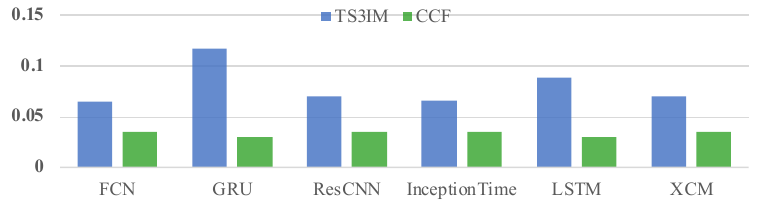} 
\caption{Bar chart illustrating the detection results of TS3IM and CCF on the FGSM-attacked samples from NonInvasiveFetalECGThorax1.}
\label{fig-fgsm}
\end{figure}

As shown in Table \ref{table_attack}, TS3IM consistently identifies differences between attacked and original samples at a ratio of approximately 2:1 compared to CCF in the majority of experiments. In specific experiments, this ratio can even reach 3:1 or higher, demonstrating TS3IM's ability to detect subtle differences even in adversarial settings. Specifically, Fig. \ref{fig-fgsm} highlights the superior detection capabilities of TS3IM on FGSM attack samples from NonInvasiveFetalECGThorax1 compared to CCF.
In individual experiments, such as where the LSTM model is subjected to DeepFool attacks on the NonInvasiveFetalECGThorax2 dataset, DTW detects a difference distance of only 0.93 and TS3IM's difference assessment is 0.0004. In contrast, CCF yields a significantly higher assessment score of 3.0545. This difference indicates that the results identified by TS3IM and DTW show a better correlation than CCF. On the same dataset where DeepFool attacks the FCN model, though the DTW result was only 66.18, the ED result in the Euclidean space reached 8251, indicating that there is a huge difference between the sequences. TS3IM also detected this difference and gave a difference score as high as 3.108. 


\section{CONCLUSIONS}

The development of the Structured TS3IM addresses a significant gap in time series analysis by introducing a comprehensive and multidimensional similarity metric. By drawing on the principles that underpin the SSIM in image quality assessment, TS3IM brings a nuanced approach to evaluating time series data, overcoming the limitations of traditional metrics. Its effectiveness and sensitivity to extreme values, demonstrated through rigorous validation and application across diverse datasets, establish TS3IM as a superior method for measuring similarity and detecting adversarial samples. This advancement not only enhances the analytical capabilities within the field of time series analysis, but also sets a new benchmark for future research and method development of data security and integrity in sensitive applications, paving the way for more accurate, secure, and insightful analysis of temporal data.

\addtolength{\textheight}{-12cm}   





\bibliographystyle{IEEEtran}

\end{document}